\documentclass[11pt]{article}
\usepackage{eacl2009}
\usepackage{times}
\usepackage{url}
\usepackage{latexsym}
\usepackage{avm}

\usepackage{amssymb,amsmath,epsfig}

\usepackage{graphicx}

\newcommand{\match}{\textsc{match}}
\newcommand{\sparky}{\textsc{sp}a\textsc{rk}y}
\newcommand{\paradise}{\textsc{paradise }}

\newcommand{\ignore}[1]{}

\newcommand{\rec}{\textsc{recommend}}
\newcommand{\comp}{\textsc{compare}}
\newcommand{\summary}{\textsc{summary}}

\title{Natural Language Generation as Planning Under Uncertainty for Spoken Dialogue Systems}

\author{
Verena Rieser\\
  School of Informatics\\
  University of Edinburgh\\
  {\tt vrieser@inf.ed.ac.uk}
 \And
  Oliver Lemon\\
 School of Informatics\\
  University of Edinburgh\\
  {\tt olemon@inf.ed.ac.uk}
  }

\date{}

\begin{document}
\maketitle

\begin{abstract}
We present and evaluate a new model for Natural Language Generation
(NLG) in Spoken Dialogue Systems, based on statistical planning, given
noisy feedback from the current generation context (e.g.\ a user and a
surface realiser).  We study its use in a standard NLG problem: how to
present information (in this case a set of search results) to users,
given the complex trade-offs between utterance length, amount of
information conveyed, and cognitive load. We set these trade-offs by
analysing existing \match{} data. 
We then train a NLG policy using Reinforcement Learning (RL), which adapts
its behaviour to noisy feedback from the current generation context.
This policy is compared to several baselines derived from previous
work in this area. The learned policy significantly outperforms all
the prior approaches.
\end{abstract}

\section{Introduction}\label{sec:intro}

Natural language allows us to achieve the same communicative goal
(``what to say") using many different expressions (``how to say it").
In a Spoken Dialogue System (SDS), an abstract communicative goal (CG) can be generated in many different ways. For example, the CG to present database results to the user can be realised as a summary  \cite{polifroni:acl08,demberg:eacl06}, or by comparing items \ \cite{match}, or by picking one item and recommending it to the user \cite{yswy07}.

\ignore{
In a Spoken Dialogue System (SDS), for example, the abstract communicative
goal {\it askName}, as defined by the Dialogue Manager, can be
generated as {\it ``Sorry, name please?"} or {\it ``First and last
  name?"}  \cite{williams:icslp08} -- one or the other might be
appropriate dependent on the the current dialogue context.
}

Previous work has shown that it is useful to adapt the generated
output to certain features of the dialogue context, for example user
preferences, e.g.\ \cite{match,demberg:eacl06}, user knowledge, e.g.\
\cite{srini:londial08}, or predicted TTS quality, e.g.\
\cite{nakatsu:acl06}.

In extending this previous work we treat NLG as a statistical sequential
planning problem, analogously to current statistical approaches to
Dialogue Management (DM), e.g.\ \cite{njfun,talk:cl08,rl-nle07} and ``conversation
as action under uncertainty" \cite{paek:AI00}.
\ignore{
In extending this previous work we treat NLG as statistical sequential
planning problem, analogously to current statistical approaches to
Dialogue Management (DM), e.g.\ \cite{talk:cl08,rl-nle07}.}
In NLG we have similar trade-offs and unpredictability as in DM, and in some systems the content planning and DM tasks are overlapping.
Clearly, very long system utterances with many actions in
them are to be avoided, because users may become confused or
impatient, but each individual NLG action will convey some
(potentially) useful information to the user. There is therefore an
optimisation problem to be solved. Moreover, the user judgements or
next (most likely) action after each NLG action are unpredictable, and
the behaviour of the surface realiser may also be variable (see
Section \ref{ssec:nlg}).

NLG could therefore fruitfully be approached as a sequential statistical
planning task, where there are trade-offs and decisions to make, such
as whether to choose another NLG action (and which one to choose) or
to instead stop generating.  Reinforcement Learning (RL) allows us to optimise such trade-offs
in the presence of uncertainty, i.e.\ the chances of achieving a
better state, while engaging in the risk of choosing another action.

In this paper we present and evaluate a new model for NLG in Spoken Dialogue
Systems as planning under uncertainty. 
 In
Section \ref{sec:motivation} we argue for applying RL to NLG problems
and explain the overall framework.  In Section \ref{sec:task} we
discuss challenges for NLG for Information Presentation.  In Section
\ref{sec:data} we present results from our analysis of the \match{}
corpus \cite{match}.  
 In Section \ref{sec:method} 
we present a detailed example of our proposed NLG method.  In Section
\ref{sec:exp} we report on experimental results using this framework
for exploring Information Presentation policies.  In Section
\ref{sec:con} we conclude and discuss future directions.

\ignore{
We apply this framework to content planning for Information
Presentation, where there are complex trade-offs and decisions to make
such as whether to choose another NLG action (and which one to choose)
or to stop generating.

In Section \ref{sec:motivation} we describe the overall framework of
applying RL to NLG problems and explain why RL is well suited to
address them.  In Section \ref{sec:task} we describe the Information
Presentation task in general.  
In Section \ref{sec:data} we present
results from our analysis of the \match{} corpus \cite{match}.  In
Section \ref{sec:method} we present a detailed example of our proposed
NLG method.  In Section \ref{sec:exp} we report on experimental
results using this framework 
for exploring Information Presentation policies.  In Section
\ref{sec:con} we conclude and discuss future directions.
}

\section{NLG as planning under uncertainty}\label{sec:motivation}

We adopt the general framework  of NLG as planning under
uncertainty (see \cite{lemon:londial08} for the initial version of this approach). 
 Some aspects of NLG
have been treated as planning, e.g.\ \cite{koller07,Koller-Petrick:icasp08}, but never before as
statistical planning. 
%

NLG actions take place in a stochastic environment, for example
consisting of a user, a realizer, and a TTS system, where the
individual NLG actions have uncertain effects on the environment. For
example, presenting differing numbers of attributes to the user, and
making the user more or less likely to choose an item, as shown by
\cite{rl:acl08} for multi-modal interaction.

Most SDS employ fixed template-based generation. Our goal, however, is to
employ a stochastic realizer for SDS, see for example \cite{sparky}.
This will introduce additional noise, which higher level NLG decisions
will need to react to.
In our framework, the NLG component must achieve a high-level
Communicative Goal from the Dialogue Manager (e.g. to present a number of items) 
through planning a sequence of
 lower-level generation steps or actions, for
example first to summarise all the items and then to recommend the highest ranking one.
Each such action has unpredictable effects due to the stochastic realizer.  For
example the realizer might employ 6 attributes when recommending item
$i_4$, but it might use only 2 (e.g.\ price and cuisine for
restaurants), depending on its own processing constraints (see e.g.\
the realizer used to collect the \match{} project data).  Likewise,
the user may be likely to choose an item after hearing a summary, or
they may wish to hear more. Generating appropriate language in context
(e.g.\ attributes presented so far) thus has the following important
features in general:
\begin{itemize}
\itemsep=0.2\itemsep
\parsep=0.2\parsep
\item NLG is {\it goal driven} behaviour
\item NLG must plan a {\it sequence} of actions
\item each action {\it changes} the environment state or context
\item the effect of each action is {\it uncertain}.
\end{itemize}

These facts make it clear that the problem of planning how to generate an utterance falls naturally into the class of statistical planning problems, rather than rule-based approaches such as 
\cite{flights04,match},  or supervised learning as explored in previous work, such as classifier
learning and re-ranking, e.g.\ \cite{sparky,oh}.
 Supervised approaches  involve the ranking of a set of completed plans/utterances and as such cannot adapt online to the context or the user.   
Reinforcement Learning (RL) provides a principled, data-driven optimisation
framework for our type of  planning problem \cite{sutton:98}. 


 \ignore{

\subsection{Analogy with Dialogue Management}

In recent years statistical planning techniques have been recognised
as an important approach to the problems of Dialogue Management (DM)
\cite{njfun,yswy07,talk:cl08}.  In DM there is a standard trade-off
between dialogue length and task success.  Longer dialogues increase
the chance of task success (for example getting the required
information from a user), but longer dialogues are usually rated more
poorly, and success depends on the user's actions, which are
unpredictable.

In NLG we have similar trade-offs and unpredictability. Clearly, very
long utterances with many actions in them are to be avoided, because
users may become confused or impatient, but each individual NLG action
will convey some (potentially) useful information to the user. There
is therefore an optimisation problem to be solved. Moreover, the user
judgements or next (most likely) action after each NLG action are
unpredictable, and the behaviour of the surface realizer may also be
variable (see example above).

NLG can therefore fruitfully be approached as a sequential statistical
planning task, where there are trade-offs and decisions to make, such
as whether to choose another NLG action (and which one to choose) or
to instead stop generating.  RL allows us to optimise such trade-offs
in the presence of uncertainty, i.e.\ the chances of achieving a
better state, while engaging in the risk of choosing another action.

}


\begin{table*}[htdp]
\begin{center}
\begin{tabular}{|l|p{8cm}|c|c|}
\hline
strategy &  example &$ av. \#attr$ & $av. \#sentence$ \\
\hline \hline
 \summary & {\it ``The 4 restaurants differ in food quality, and cost."} ($\#attr=2, \#sentence=1$)& 2.07$\pm$.63  & 1.56$\pm$.5 \\
 \hline
\comp & {\it``Among the selected restaurants, the following offer exceptional overall value. Aureole's price is 71 dollars. It has superb food quality, superb service and superb decor. Daniel's price is 82 dollars. It has superb food quality, superb service and superb decor." } ($\#attr=4, \#sentence=5$)& 3.2$\pm$1.5 & 5.5$\pm$3.11 \\
\hline
 \rec  &{\it``Le Madeleine has the best overall value among the selected restaurants. Le Madeleine's price is 40 dollars and It has very good food quality. It's in Midtown West. "} ($\#attr=3, \#sentence=3$) & 2.4$\pm$.7 & 3.5$\pm$.53  \\
\hline

\end{tabular}
\caption{NLG strategies present in the \match{} corpus with average no. attributes and sentences as found in the data.}
\end{center}
\label{tab:match-strategies}
\end{table*}%


\section{The Information Presentation Problem}\label{sec:task}

We will tackle the well-studied problem of Information Presentation in NLG  to show the benefits of this approach.
The task here is to find the best way to present a set of search
results to a user (e.g.\ some restaurants meeting a certain set of
constraints). This is a task common to much prior work in NLG, e.g.\
\cite{match,demberg:eacl06,polifroni:acl08}.

In this problem, there there are many decisions  
 available for exploration.  For instance, which presentation strategy to apply ({\em NLG strategy selection}),
 how many attributes of each item to present ({\em attribute selection}),
 how to rank the items and attributes according to different models of
 user preferences ({\em attribute ordering}), how many (specific)
 items to tell them about ({\em conciseness}), how many sentences to
 use when doing so ({\em syntactic planning}), and which words to use ({\em lexical choice}) etc.  All these parameters
 (and potentially many more) can be varied, and ideally, jointly
 optimised based on  user judgements.

We had two corpora available to study some of the regions of this
decision space.  We utilise the \match{} corpus \cite{match} to
extract an evaluation function (also known as ``reward function'') for RL.
Furthermore, we utilise the \sparky{} corpus \cite{sparky} to build a
high quality stochastic realizer.  Both corpora contain data from
``overhearer" experiments targeted to Information Presentation in
dialogues in the restaurant domain.  While we are ultimately
interested in how hearers {\it engaged} in dialogues judge different Information Presentations,
results from overhearers are still directly relevant to the task.


\section{\match{} corpus analysis}\label{sec:data}

The \match{} project  made two data sets available, see
\cite{Stent:icslp02} and \cite{whittaker:eurospeech03}, which we
combine to define an evaluation function for different Information
Presentation strategies.

The first data set, see \cite{Stent:icslp02}, comprises 1024 ratings
by 16 subjects (where we only use the speech-based half, $n=512$)
on the following presentation strategies: \rec,
\comp, \summary. These strategies are realised using templates as in Table \ref{tab:match-strategies}, and varying numbers of attributes. 
In this study the users rate the individual presentation strategies as significantly different
($F(2)=1361, p<.001$). We find that \summary{} is rated significantly
worse ($p=.05$ with Bonferroni correction) than \rec{} and \comp,
which are rated as equally good.

This suggests that one should never generate a \summary. However, \summary{} has different qualities from \comp{} and \rec, as it gives users a general overview of the domain, and probably  helps the
user  to feel more confident when choosing an item, especially when they are unfamiliar with the domain,   as shown by \cite{polifroni:acl08}.

In order to further describe the strategies, we extracted different
surface features as present in the data (e.g.\ number of attributes
realised, number of sentences, number of words, number of database
items talked about, etc.) and performed a stepwise linear regression
 to find the features which were important to the overhearers
(following the \paradise framework \cite{paradise}). 
 We discovered  a trade-off between the {\em length} of the utterance ($\#sentence$) and the number of
attributes realised ($\#attr$), i.e.\ its {\em informativeness}, where overhearers like to hear as many
attributes as possible in the most concise way, as indicated by the regression
model shown in Equation \ref{eq:regression} ($R^2=.34$).~\footnote{For comparison: \cite{paradise} report on $R^2$ between .4 and .5 on a slightly lager data set.} 

\vspace{-0.5cm}
\begin{equation}\label{eq:regression}
score = .775\times\#attr + (-.301)\times\#sentence;
\end{equation}
\vspace{-0.5cm}

\begin{figure*}[htbp!]
\begin{center} 
\includegraphics[scale=0.5]{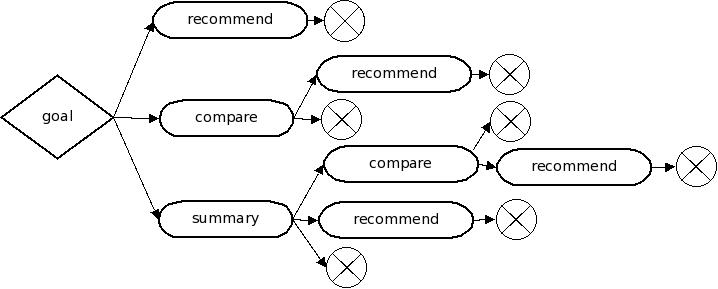} 
\caption{Possible NLG policies (X=stop generation)}\label{fig:policies}
\end{center} 
\end{figure*}

\ignore{
\begin{table*}[ht!]
\begin{center}
\begin{tabular}{|p{17cm}|}
\hline
{\em  115 restaurants meet your query.  There are 82 restaurants  which are in the cheap price range and have good or excellent food quality.  16 of them are Italian, while   12 of them are Scottish,   9 of them are international and 45 have different types of cuisine. There are also 33 others which have poor food quality.  Black Bo's and Gordon's Trattoria  are both in the cheap price range. Black Bo's has excellent food quality, while Gordon's Trattoria has good food quality.  Black Bo's is a Vegetarian restaurant. This restaurant is located in Old Town. Black Bo's has the best overall quality amongst the selected restaurants.}\\
\hline
\end{tabular}
\end{center}
\label{tab:ex-sequence}
\caption{Example utterance presenting information as a sequence: \summary{} generated as associative clustering with user modelling and 3 attributes; \comp{} 2 restaurants by 2 attributes; and \rec{} with 2 attributes and the claim last. }
\end{table*}
}

The second \match{} data set, see \cite{whittaker:eurospeech03},
comprises 1224 ratings by 17 subjects on the NLG strategies \rec{} and
\comp. The strategies realise varying numbers of attributes according
to different ``conciseness" values: {\tt concise} (1 or 2 attributes),
{\tt average} (3 or 4), and {\tt verbose} (4, 5, or 6). Overhearers
rate all conciseness levels as significantly different
($F(2)=198.3,p<.001$), with {\tt verbose} rated highest and {\tt
  concise} rated lowest, supporting our findings in the first data
set.  However, the relation between number of attributes and user
ratings is not strictly linear: ratings drop for $\#attr=6$.  This
suggests that there is an upper limit on how many attributes users
like to hear. We expect this to be especially true for real users
engaged in actual dialogue interaction, see \cite{winterboer:is07}. We
therefore  include ``cognitive load" as a variable when training
the policy (see Section \ref{sec:exp}).

In addition to the trade-off between {\em length} and {\em informativeness} for single NLG strategies, we are interested whether this trade-off will also hold for generating {\em sequences} of NLG actions.
 \cite{whittaker:lrec02}, for example, generate a {\it combined strategy} where first a  \summary{} is used to describe the retrieved subset and then they \rec{} one specific item/restaurant. For example {\it ``The 4 restaurants are all French, but differ in food quality, and cost. Le Madeleine has the best overall value among the selected restaurants. Le Madeleine's price is 40 dollars and It has very good food quality. It's in Midtown West."} 

We therefore extend the set of possible strategies present in the data
for exploration: we allow ordered combinations of the strategies,
assuming that only \comp{} or \rec{} can follow a \summary, 
and that only \rec{} can follow \comp, resulting in 7 possible actions: 

\begin{enumerate}
\itemsep=0.2\itemsep
\parsep=0.2\parsep
\item \rec
\item \comp
\item \summary
\item \comp+\rec
\item \summary+\rec
\item \summary+\comp
\item \summary+\comp+\rec
\end{enumerate}

We then analytically solved the
regression model in Equation \ref{eq:regression} for the 7 possible
strategies using average values from the \match{} data. This is solved by a system of linear
inequalities. 
According to this model, the best ranking strategy is
to do all the presentation strategies in one sequence, i.e.\
\summary+\comp+\rec. However, this analytic solution assumes a
``one-shot" generation strategy where there is no intermediate
feedback from the environment: users are simply static overhearers
(they cannot ``barge-in'' for example), there is no variation in
the behaviour of the surface realizer, i.e.\ one would use fixed templates as in \match, and the user has unlimited cognitive capabilities. 
These assumptions are not realistic, and must be relaxed. 
%
In the next Section we describe a worked through example of the overall framework. 



\begin{figure*}[ht!]
\begin{center}
\includegraphics[scale=0.25]{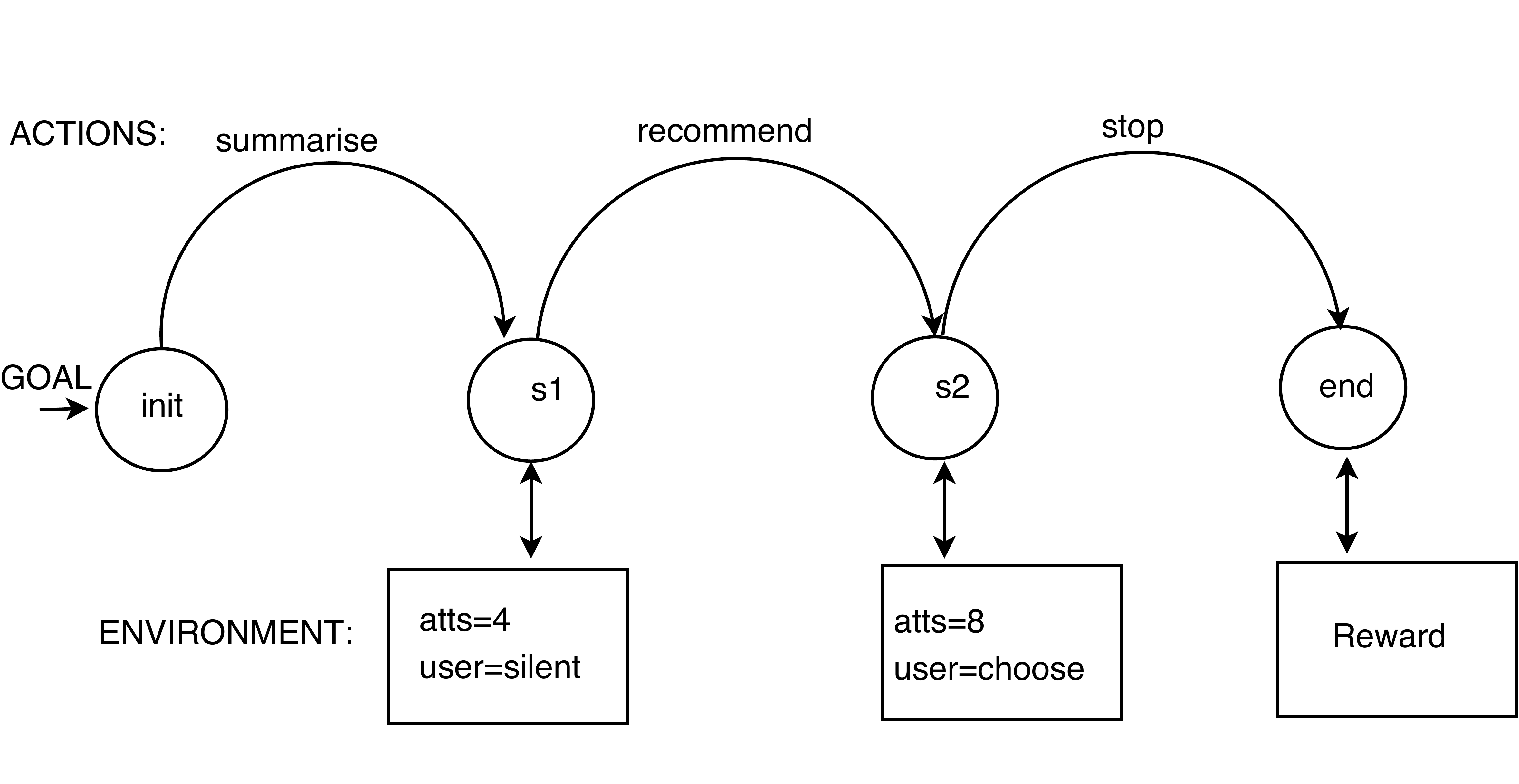} 
\caption{Example RL-NLG action sequence for Table  \ref{tab:exampleplan}}
\label{fig:exampleplan}
\end{center} 
\end{figure*}

\begin{table*}[ht!]
\hspace{-1.6cm}
\begin{tabular}{|p{0.5cm}|l|c|}
\hline
{\bf State}& {\bf Action} & {\bf State change/effect} \\
\hline
init& SysGoal: {\tt present\_items($i_1,i_2,i_5,i_8$)\& user\_choose\_one\_of($i_1,i_2,i_5,i_8$)} & initialise state\\

\hline
s1& RL-NLG: \summary($i_1,i_2,i_5,i_8$) &  att=4, sent=1, user=silent\\   
s2& RL-NLG: \rec($i_5$)& att=8, sent=4, user=choose($i_5$)\\
end & RL-NLG: stop   &  calculate Reward\\

\hline
\end{tabular}
\label{tab:exampleplan}
\caption{Example utterance planning sequence for Figure \ref{fig:exampleplan}}
\end{table*}


\section{Method: the RL-NLG model}\label{sec:method}

For the reasons discussed above, we treat the NLG module as a
statistical planner, operating in a stochastic environment, and
optimise it using Reinforcement Learning.  The input to the module is
a Communicative Goal  supplied by the Dialogue Manager. The CG
consists of a Dialogue Act to be generated, for example {\tt
present\_items($i_1,i_2,i_5,i_8$)}, and a System Goal ({\tt
SysGoal}) which is the desired user reaction, e.g.\ to make the user
choose one of the presented items ({\tt
user\_choose\_one\_of($i_1,i_2,i_5,i_8$})). The RL-NLG module must
plan a sequence of lower-level NLG actions that achieve the goal (at
lowest cost) in the current context. The context consists of a user
(who may remain silent, supply more constraints, choose an item, or
quit), and variation from the sentence realizer described above.

Now let us walk-through one simple utterance plan as carried out by
this model, as shown in Table  2. 
 Here, we start with the CG
{\tt present\_items($i_1,i_2,i_5,i_8$)\&
user\_choose\_one\_of($i_1,i_2,i_5,i_8$)} from the system's
DM. This initialises the NLG state ($init$). The policy
chooses the action \summary{} 
 and this transitions us to state
$s1$, where we observe that 4 attributes and 1 sentence have been
generated, and the user is predicted to remain silent. In this state, the current NLG
policy is to \rec{} the top ranked item ($i_5$, for this user),
which takes us to state $s2$, where 8 attributes have been generated
in a total of 4 sentences, and the user chooses an item. 
The policy holds that in states like $s2$ the best thing to do is
``stop'' and pass the turn to the user. This takes us to the state
$end$, where the total reward of this action sequence is computed (see
Section \ref{ssect:reward}), and used to update the NLG policy in each of
the visited state-action pairs via back-propagation.


\section{Experiments}\label{sec:exp}

We now report on a proof-of-concept study where we train our policy in a simulated learning environment based on the results from the \match{} corpus analysis in Section \ref{sec:data}. 
Simulation-based RL allows to explore unseen actions which are not in the data, and thus less initial data is needed \cite{rl:acl08}.
Note, that we cannot directly learn from the \match{} data, as therefore we would need data from an interactive dialogue. We are currently collecting such data in a Wizard-of-Oz experiment.

\subsection{User simulation}\label{ssect:}

User simulations are commonly used to train strategies for Dialogue
Management, see for example \ \cite{yswy07}. A user
simulation for NLG is very similar, in that it is a predictive model
of the most likely next user act. However, this user act does not
actually change the overall dialogue state (e.g.\ by filling slots)
but it only changes the generator state. In other words, the NLG user
simulation tells us what the user is most likely to do next, {\em if we
were to stop generating now}. It also tells us the probability whether
the user chooses to ``barge-in'' after a system NLG action (by either
choosing an item or providing more information).

The user simulation for this study is a simple bi-gram model, 
which relates the number of attributes presented to
the next likely user actions, see Table \ref{tab:user-sim}. The user
can either follow the goal provided by the DM ({\tt SysGoal}), for
example  choosing an item. The user can also do something else ({\tt
userElse}), e.g.\ providing another constraint, or the user can quit ({\tt
userQuit}).

For simplification, we discretise the number of attributes into {\tt
concise-average-verbose}, reflecting the conciseness values from the \match{} data, as described
in Section \ref{sec:data}.
 In addition, we assume that the user's
cognitive abilities are limited (``cognitive load"), based on the results from the second \match{} data set in Section \ref{sec:data}. Once the number of attributes is more than the ``magic number 7" (reflecting psychological results
on short-term memory) \cite{baddeley:03}
  the user is more likely to become
confused and quit.

The probabilities in Table \ref{tab:user-sim} are currently manually
set heuristics. 
  We are currently conducting  a Wizard-of-Oz study in order to learn
these probabilities (and other user parameters) from real data.

\begin{table}[htdp]
\small
\begin{center}
\begin{tabular}{|p{1.5cm}|p{1.5cm}p{1.6cm}p{1.6cm}|}
\hline
     & {\tt SysGoal} & {\tt userElse} & {\tt userQuit} \\
 \hline
{\tt concise} &  20.0 & 60.0  & 20.0 \\
{\tt average} &  60.0 & 20.0  & 20.0 \\
{\tt verbose} &  20.0 & 20.0  & 60.0 \\
\hline
\end{tabular}
\end{center}
\caption{NLG bi-gram user simulation}
\label{tab:user-sim}
\end{table}%

\subsection{Realizer model}\label{ssec:nlg}

The sequential NLG model assumes a realizer, which updates the context after each generation
step (i.e.\ after each single action). We estimate the realiser's parameters from the mean values we
found in the \match{} data (see Table 1). 
For this study we first (randomly) vary the number of
attributes, whereas the number of sentences is fixed (see Table
\ref{tab:realizer}). 
 In current work we replace the realizer model with an implemented generator that replicates the variation found in the \sparky{} realizer \cite{sparky}. 

\begin{table}[htdp]

\begin{center}
\begin{tabular}{|l|cc|}
\hline
 & $\#attr$ & $\#sentence$\\
 \hline
\summary & 1 or 2 & 2\\
\comp & 3 or 4 & 6 \\
\rec & 2 or 3 & 3 \\
\hline
\end{tabular}
\end{center}
\caption{Realizer parameters}
\label{tab:realizer}
\end{table}%

  \begin{figure*}[htbp!]
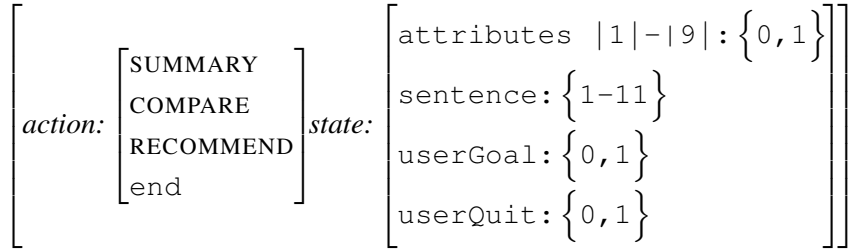

\begin{avm}
\[
{\sc \em action:} \[{\summary} \\
              {\comp}\\
              {\rec}\\
              {\tt end} 
                \]
                {\sc \em state:} \[{\tt attributes \|1\|-|9\|:} \{{\tt 0,1}\}\\
              {\tt sentence:} \{{\tt 1-11}\}\\
              {\tt userGoal:} \{{\tt 0,1}\}\\
               {\tt userQuit:} \{{\tt 0,1}\}
                \]
           \]
\end{avm} 
\normalsize
\centering \caption{State-Action space for RL-NLG} \label{fig:state-action-space}
\end{figure*}

\subsection{Reward function}\label{ssect:reward}

The reward function defines the final goal of the
utterance generation sequence.  In this experiment the reward is a
function of the various data-driven trade-offs as identified in the
data analysis in Section
\ref{sec:data}: utterance length and number of provided attributes, as
weighted by the regression model in Equation \ref{eq:regression}, as
well as the next predicted user action. 
Since we currently only have overhearer data, we manually estimate the reward for the next most likely user act, to supplement the data-driven model.
If in the
{\it end} state the next most likely user act is {\tt userQuit}, the
learner gets a penalty of $-100$, {\tt userElse} receives 0
reward, and {\tt SysGoal} gains $+100$ reward. Again, these 
hand coded scores need to be refined by a more targeted data collection, but the other components of the reward function are data-driven.

Note that RL learns to ``make compromises" with respect to the different trade-offs. For example, the user is less likely to choose an item if there are more than 7 attributes, but the realizer can generate 9 attributes. However, in some contexts it might be desirable to generate all 9 attributes, e.g.\ if the generated utterance is short.  
Threshold-based approaches, in contrast, cannot (easily) reason with respect to the current context.

\subsection{State and Action Space}\label{ssect:}

We now formulate the problem as a Markov Decision Process (MDP),
relating states to actions. 
Each state-action pair is associated with a {\em transition
probability}, which is the probability of moving from state $s$ at
time $t$ to state $s'$ at time $t+1$ after having performed action $a$
when in state $s$. This transition probability is computed by the
environment model (i.e.\ user and realizer), and
explicitly captures noise/uncertainty in the environment. This is a
major difference to other non-statistical planning approaches.
Each transition is also associated with a
reinforcement signal (or reward) $r_{t+1}$ describing how good the
result of action $a$ was when performed in state $s$.

The state space comprises 9 binary features representing the number of attributes, 2 binary features representing the predicted user's next action to follow the system goal or quit,
as well as a discrete feature reflecting the number of sentences generated so far, as shown in Figure \ref{fig:state-action-space}.
This results in \underline{$2^{11} \times 6 = 12,288$} distinct generation states. We trained the policy using the well known \textsc{sarsa} algorithm, using linear function approximation \cite{sutton:98}.
The policy was trained for 3600 simulated NLG sequences.

In future work we plan to learn lower level decisions, such as lexical adaptation based on the vocabulary used by the user.

\subsection{Baselines}\label{ssect:}

We derive the baseline policies from Information Presentation
strategies as deployed by current dialogue systems. In total we
utilise 7 different baselines (B1-B7), which correspond to single branches in our policy space (see Figure \ref{fig:policies}):

\begin{description}
\itemsep=0.1\itemsep
\parsep=0.1\parsep
\item[B1:]  \rec{} only, e.g.\ \cite{yswy07}
\item[B2:]  \comp{} only, e.g.\  \cite{talk:cl08} 
\item[B3:]  \summary{} only, e.g.\  \cite{polifroni:acl08}
\item[B4:] \summary{} followed by \rec, e.g.\  \cite{whittaker:lrec02}
\item[B5:]  Randomly choosing between \comp{} and \rec, e.g.\  \cite{walker:jair07}
\item[B6:]  Randomly choosing between all 7 outputs
\item[B7:]  Always generating whole sequence, i.e.\ \summary+\comp+\rec, as suggested by the analytic solution  (see Section \ref{sec:data}).
\end{description}

\subsection{Results}\label{ssect:}

We analyse the test runs (n=200) using an ANOVA with a Post-Hoc T-Test
(with Bonferroni correction). RL significantly ($p<.001$) outperforms
all baselines in terms of final reward, see Table
\ref{tab:results}. 
RL is the only policy which significantly
improves the next most likely user action by adapting to features in the current context. In contrast to conventional approaches,  RL learns to `control' its environment according to the estimated transition probabilities and the associated rewards.

The learnt policy can be described as follows: It either starts with
\summary{} or \comp{} after the {\it init} state, i.e.\ it learnt to never start with a \rec. 
It stops generating
after \comp{} if the {\tt userGoal} is (probably) reached (e.g.\ the user is most likely to choose an item in the next turn, which depends
on the number of attributes generated), otherwise it goes on and
generates a \rec.  If it starts with \summary, it always generates a
\comp{} afterwards. Again, it stops if the {\tt userGoal} is (probably) reached,
otherwise it generates the full sequence (which corresponds to the
analytic solution B7).

The analytic solution B7 performs second best, and significantly
outperforms all the other baselines ($p<.01$). Still, it is
significantly worse ($p<.001$) than the learnt policy as this
`one-shot-strategy' cannot robustly and dynamically adopt to noise or
changes in the environment.

In general, generating sequences of NLG actions rates higher than
generating single actions only: B4 and B6 rate directly after RL and B7, while
B1, B2, B3, B5 are all equally bad given our data-driven definition of
reward and environment. Furthermore, the simulated environment allows us to replicate the results in the \match{} corpus (see Section \ref{sec:data}) when only comparing single strategies: \summary{} performs significantly worse, while \rec{} and \comp{} perform equally well.

\begin{table}[htdp]
\begin{center}
\begin{tabular}{|l|ll|}
\hline
policy & reward & ($\pm std$)\\
\hline 
B1	&	99.1	& ($\pm$129.6) \\
B2	&	90.9	&($\pm$142.2) \\
B3	&	65.5	&($\pm$137.3)\\
B4	&	176.0&($\pm$154.1)\\
B5	&	95.9	 &($\pm$144.9)\\
B6	&	168.8 &($\pm$165.3)\\
B7	&	229.3 &($\pm$157.1)\\
RL	&	310.8 &($\pm$136.1)\\
\hline
\end{tabular}
\end{center}
\caption{Evaluation Results ($p<.001$ ) }
\label{tab:results}
\end{table}%


\section{Conclusion}\label{sec:con}

We presented and evaluated a new model for Natural Language
 Generation (NLG) in Spoken Dialogue Systems, based on statistical
 planning.  After motivating and presenting the model, we studied its
 use in  Information Presentation.

We derived a data-driven model predicting users' judgements
on different information presentation actions (reward function), via a regression
analysis on \match{}  data. We used this regression model to
set weights in a reward function for Reinforcement Learning, and so
optimise a context-adaptive presentation policy.
 The learnt policy was compared to several baselines derived from previous
 work in this area, where the learnt policy significantly outperforms all
 the baselines.  
 
 There are many possible extensions to this
 model, e.g.\ using the same techniques to jointly optimise choosing the
 number of attributes, aggregation, word choice, referring
 expressions, and so on, in a hierarchical manner.

We are currently collecting data in  targeted Wizard-of-Oz
experiments, to derive a fully data-driven training
environment and test the learnt policy with real users, following \cite{rl:acl08}. 
The trained NLG strategy will also be integrated in an end-to-end statistical system within the \textsc{class}i\textsc{c} project (\url{www.classic-project.org}).



\section*{Acknowledgments}
We thank Marilyn Walker for access to the   \match{} corpus.
	The research leading to these results has received funding from the European Community's Seventh Framework Programme (FP7/2007-2013) under grant agreement no. 216594 (\textsc{class}i\textsc{c} project project: \url{www.classic-project.org}) and from the EPSRC project no. EP/E019501/1.
\normalsize

\bibliographystyle{acl}
\bibliography{eacl09}

\begin{thebibliography}{}

\bibitem[\protect\citename{Baddeley}2001]{baddeley:03}
A.~Baddeley.
\newblock 2001.
\newblock Working memory and language: an overview.
\newblock {\em Journal of Communication Disorder}, 36(3):189--208.

\bibitem[\protect\citename{Demberg and Moore}2006]{demberg:eacl06}
Vera Demberg and Johanna~D. Moore.
\newblock 2006.
\newblock Information presentation in spoken dialogue systems.
\newblock In {\em Proceedings of EACL}.

\bibitem[\protect\citename{Henderson \bgroup et al.\egroup }2008]{talk:cl08}
James Henderson, Oliver Lemon, and Kallirroi Georgila.
\newblock 2008.
\newblock Hybrid reinforcement / supervised learning of dialogue policies from
  fixed datasets.
\newblock {\em Computational Linguistics}, 34:4.

\bibitem[\protect\citename{Janarthanam and Lemon}2008]{srini:londial08}
Srinivasan Janarthanam and Oliver Lemon.
\newblock 2008.
\newblock {User simulations for online adaptation and knowledge-alignment in
  Troubleshooting dialogue systems}.
\newblock In {\em Proc. of SEMdial}.

\bibitem[\protect\citename{Koller and Petrick}2008]{Koller-Petrick:icasp08}
Alexander Koller and Ronald Petrick.
\newblock 2008.
\newblock Experiences with planning for natural language generation.
\newblock In {\em ICAPS}.

\bibitem[\protect\citename{Koller and Stone}2007]{koller07}
Alexander Koller and Matthew Stone.
\newblock 2007.
\newblock Sentence generation as planning.
\newblock In {\em Proceedings of ACL}.

\bibitem[\protect\citename{Lemon}2008]{lemon:londial08}
Oliver Lemon.
\newblock 2008.
\newblock {Adaptive Natural Language Generation in Dialogue using Reinforcement
  Learning}.
\newblock In {\em Proceedings of SEMdial}.

\bibitem[\protect\citename{Moore \bgroup et al.\egroup }2004]{flights04}
Johanna Moore, Mary~Ellen Foster, Oliver Lemon, and Michael White.
\newblock 2004.
\newblock Generating tailored, comparative descriptions in spoken dialogue.
\newblock In {\em Proc.\ FLAIRS}.

\bibitem[\protect\citename{Nakatsu and White}2006]{nakatsu:acl06}
Crystal Nakatsu and Michael White.
\newblock 2006.
\newblock Learning to say it well: Reranking realizations by predicted
  synthesis quality.
\newblock In {\em Proceedings of ACL}.

\bibitem[\protect\citename{Oh and Rudnicky}2002]{oh}
Alice Oh and Alexander Rudnicky.
\newblock 2002.
\newblock Stochastic natural language generation for spoken dialog systems.
\newblock {\em Computer, Speech \& Language}, 16(3/4):387--407.

\bibitem[\protect\citename{Paek and Horvitz}2000]{paek:AI00}
Tim Paek and Eric Horvitz.
\newblock 2000.
\newblock Conversation as action under uncertainty.
\newblock In {\em Proc. of the 16th Conference on Uncertainty in Artificial
  Intelligence}.

\bibitem[\protect\citename{Polifroni and Walker}2008]{polifroni:acl08}
Joseph Polifroni and Marilyn Walker.
\newblock 2008.
\newblock {Intensional Summaries as Cooperative Responses in Dialogue
  Automation and Evaluation}.
\newblock In {\em Proceedings of ACL}.

\bibitem[\protect\citename{Rieser and Lemon}2008a]{rl-nle07}
Verena Rieser and Oliver Lemon.
\newblock 2008a.
\newblock {Does this list contain what you were searching for? Learning
  adaptive dialogue strategies for Interactive Question Answering}.
\newblock {\em J.\ Natural Language Engineering}, 15(1):55--72.

\bibitem[\protect\citename{Rieser and Lemon}2008b]{rl:acl08}
Verena Rieser and Oliver Lemon.
\newblock 2008b.
\newblock {Learning Effective Multimodal Dialogue Strategies from Wizard-of-Oz
  data: Bootstrapping and Evaluation}.
\newblock In {\em Proceedings of ACL}.

\bibitem[\protect\citename{Singh \bgroup et al.\egroup }2002]{njfun}
S.~Singh, D.~Litman, M.~Kearns, and M.~Walker.
\newblock 2002.
\newblock Optimizing dialogue management with {R}einforcement {L}earning:
  Experiments with the {NJFun} system.
\newblock {\em JAIR}, 16:105--133.

\bibitem[\protect\citename{Stent \bgroup et al.\egroup }2002]{Stent:icslp02}
Amanda Stent, Marilyn Walker, Steve Whittaker, and Preetam Maloor.
\newblock 2002.
\newblock User-tailored generation for spoken dialogue: an experiment.
\newblock In {\em In Proc. of ICSLP}.

\bibitem[\protect\citename{Stent \bgroup et al.\egroup }2004]{sparky}
Amanda Stent, Rashmi Prasad, and Marilyn Walker.
\newblock 2004.
\newblock Trainable sentence planning for complex information presentation in
  spoken dialog systems.
\newblock In {\em Association for Computational Linguistics}.

\bibitem[\protect\citename{Sutton and Barto}1998]{sutton:98}
R.~Sutton and A.~Barto.
\newblock 1998.
\newblock {\em {R}einforcement {L}earning}.
\newblock MIT Press.

\bibitem[\protect\citename{Walker \bgroup et al.\egroup }2000]{paradise}
Marilyn~A. Walker, Candace~A. Kamm, and Diane~J. Litman.
\newblock 2000.
\newblock Towards developing general models of usability with {PARADISE}.
\newblock {\em Natural Language Engineering}, 6(3).

\bibitem[\protect\citename{Walker \bgroup et al.\egroup }2004]{match}
Marilyn Walker, S.~Whittaker, A.~Stent, P.~Maloor, J.~Moore, M.~Johnston, and
  G.~Vasireddy.
\newblock 2004.
\newblock User tailored generation in the match multimodal dialogue system.
\newblock {\em Cognitive Science}, 28:811--840.

\bibitem[\protect\citename{Walker \bgroup et al.\egroup }2007]{walker:jair07}
Marilyn Walker, Amanda Stent, Fran\c{c}ois Mairesse, and Rashmi Prasad.
\newblock 2007.
\newblock Individual and domain adaptation in sentence planning for dialogue.
\newblock {\em Journal of Artificial Intelligence Research (JAIR)},
  30:413--456.

\bibitem[\protect\citename{Whittaker \bgroup et al.\egroup
  }2002]{whittaker:lrec02}
Steve Whittaker, Marilyn Walker, and Johanna Moore.
\newblock 2002.
\newblock {Fish or Fowl}: A {Wizard of Oz} evaluation of dialogue strategies in
  the restaurant domain.
\newblock In {\em Proc.\ of the International Conference on Language Resources
  and Evaluation (LREC)}.

\bibitem[\protect\citename{Whittaker \bgroup et al.\egroup
  }2003]{whittaker:eurospeech03}
Stephen Whittaker, Marilyn Walker, and Preetam Maloor.
\newblock 2003.
\newblock Should i tell all? an experiment on conciseness in spoken dialogue.
\newblock In {\em Proc.\ European Conference on Speech Processing
  (EUROSPEECH)}.

\bibitem[\protect\citename{Winterboer \bgroup et al.\egroup
  }2007]{winterboer:is07}
Andi Winterboer, Jiang Hu, Johanna~D. Moore, and Clifford Nass.
\newblock 2007.
\newblock The influence of user tailoring and cognitive load on user
  performance in spoken dialogue systems.
\newblock In {\em Proc.\ of the 10th International Conference of Spoken
  Language Processing (Interspeech/ICSLP)}.

\bibitem[\protect\citename{Young \bgroup et al.\egroup }2007]{yswy07}
SJ~Young, J~Schatzmann, K~Weilhammer, and H~Ye.
\newblock 2007.
\newblock {The Hidden Information State Approach to Dialog Management}.
\newblock In {\em ICASSP 2007}.

\end{thebibliography}

\end{document}